\documentclass[lettersize,journal]{IEEEtran}
\usepackage{amsmath,amsfonts,amssymb}
\usepackage{algorithmic}
\usepackage{algorithm}
\usepackage{array}
\usepackage[caption=false,font=normalsize,labelfont=sf,textfont=sf]{subfig}
\usepackage{textcomp}
\usepackage{stfloats}
\usepackage{url}
\usepackage{verbatim}
\usepackage{booktabs}
\usepackage{graphicx}
\usepackage{cite}
\usepackage{balance}
\usepackage{placeins}
\usepackage{float}
\usepackage{multirow}
\usepackage{xcolor}

\newcommand{\best}[1]{%
  \leavevmode
  \makebox[0pt][l]{\kern-0.18pt\textbf{#1}}%
  \makebox[0pt][l]{\kern 0.18pt\textbf{#1}}%
  \textbf{#1}%
}

\newcommand{\cmark}{\textcolor{green!60!black}{\checkmark}}
\newcommand{\xmark}{\textcolor{red!75!black}{\(\times\)}}

\begin{document}

\title{Probe-VAD: Ordinal Likelihood Probing for Training-Free Video Anomaly Detection}

\author{
Jiawei Gu$^{*}$,
Qilin Zhao$^{*}$,
Tengkuo Guo,
Zhiming Zhong,
Shuangqing Zhang$^{\dagger}$,
Fan Lyu,

Fang Zhao$^{\dagger}$, Guo-Sen Xie, Caifeng Shan, \emph{Senior Member, IEEE}
\thanks{Jiawei Gu, Zhiming Zhong, Fang Zhao, Shuangqing Zhang and Caifeng Shan are with the School of Intelligence Science and Technology, Nanjing University,
Suzhou 215163, China
(e-mail: jiaweigu51@gmail.com; 251880598@smail.nju.edu.cn;
fzhao@nju.edu.cn; zsq\_cs@foxmail.com; caifeng.shan@gmail.com).}%
\thanks{Qilin Zhao is with the Faculty of Life Science and Medicine, School of Medicine and Health, Harbin Institute of Technology, Harbin 150001, China
 (e-mail: 2024112930@stu.hit.edu.cn).}%
\thanks{Tengkuo Guo is with the School of Instrumentation Science and Engineering,
Harbin Institute of Technology, Harbin 150001, China
(e-mail: 2023113221@stu.hit.edu.cn).

Fan Lyu is with the Computer Vision Center (CVC), Universitat Autònomade Barcelona (UAB), Barcelona, 08193 Spain. (e-mail: fanlyu@cvc.uab.cat).

Guo-Sen Xie is with the School of Computer Science and Engineering, Nanjing University
of Science and Technology, Nanjing 210014, China (e-mail: guosen.xie@njust.edu.cn).
}%
\thanks{ $^{*}$~Contributed equally. $^{\dagger}$~Corresponding Author.}%
}



\maketitle

\begin{abstract}
Video anomaly detection (VAD) aims to localize anomalous events in untrimmed videos. Vision-language models (VLMs) provide rich visual understanding for training-free VAD, but existing approaches impose restrictive interfaces between visual understanding and anomaly scoring. Caption-based pipelines compress visual evidence into text, potentially discarding subtle cues, while direct numerical generation forces the model to express its judgment through a small set of predefined scores. Such interfaces can obscure subtle differences in anomaly severity, causing visually distinct clips to receive similar representations or scores and thereby limiting the resolution of anomaly ranking. We propose \textbf{Probe-VAD}, an ordinal binary-probing framework that directly probes severity preferences from a frozen VLM. Given raw video clips, Probe-VAD queries ten ordered severity thresholds and extracts constrained \textit{YES}/\textit{NO} continuation likelihoods. Their normalized preferences form a cumulative severity profile, from which tail evidence is aggregated into a continuous anomaly score, with isotonic projection enforcing ordinal consistency. Experiments on public VAD benchmarks demonstrate superior performance with low computational cost. Probe-VAD provides a simple interface for translating frozen VLM visual understanding into continuous, rank-sensitive anomaly scores without task-specific training or caption-based compression. Code is available at: \url{https://github.com/yvestine/COVAS-VAD}.

\end{abstract}


\section{Introduction}
\label{sec:introduction}

Video anomaly detection (VAD)~\cite{zhang2026reconstructive, wu2026deep} aims to temporally localize unusual and hazardous events in long untrimmed videos, and has attracted significant attention due to its overcoming the impracticality of manual monitoring in surveillance security~\cite{sultani2018real}, and industrial inspection~\cite{dong2026cl, duan2025anomalycontrol}. Previous VAD approaches have achieved compelling performance under one-class~\cite{cao2024context,leng2022anomaly} and weakly supervised settings~\cite{sultani2018real,wu2021learning,wu2024vadclip,zhang2026contextual}, but typically rely on task- or domain-specific optimization. Such dependence can limit their generalization when test-time scenes, anomaly categories, or operational conditions differ from those encountered during training.
Recent advances in vision-language models (VLMs)~\cite{alayrac2022flamingo} offer a promising alternative for training-free VAD, leveraging rich pretrained visual-semantic knowledge without target-domain optimization. Accordingly, recent methods have explored caption-driven linguistic reasoning~\cite{zhang2026contextual}, event-aware and hierarchical temporal decomposition~\cite{shao2025eventvad,li2025vadtree}, direct generative refinement~\cite{lim2026corevad}, and memory-augmented retrieval~\cite{lee2025flashback}. These efforts demonstrate the strong potential of frozen VLMs for perceiving and reasoning about anomalous video content. However, rich visual understanding alone is insufficient for effective VAD, which requires translating visual evidence into continuous anomaly scores for fine-grained temporal ranking. Existing studies have largely focused on how visual evidence is represented, organized, or reasoned over, while comparatively less attention has been paid to the interface through which VLM understanding is converted into anomaly scores. This leaves a fundamental question largely underexplored: \emph{how can the rich visual understanding of a frozen VLM be faithfully translated into continuous anomaly-ranking signals?}

\begin{figure*}[!t]
	\centering
	\includegraphics[width=\linewidth]{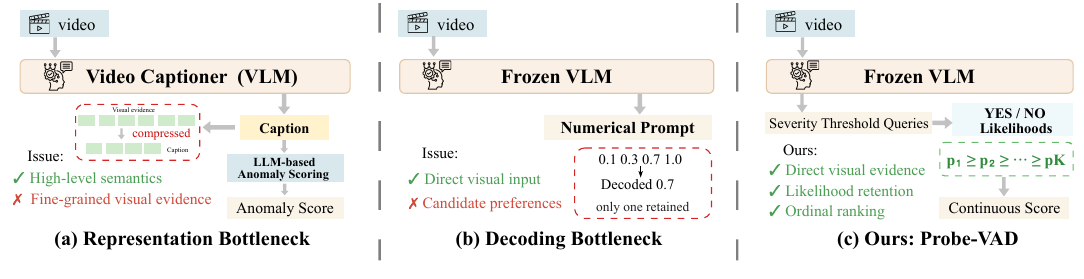}
    \caption{
(a) Caption-mediated VAD compresses visual observations into text, potentially discarding fine-grained anomaly cues before scoring and creating a \textbf{representation bottleneck}.
(b) Direct numerical VLM scoring retains visual evidence but collapses richer model preferences into a single decoded value, creating a \textbf{decoding bottleneck} that limits fine-grained anomaly ranking.
(c) Probe-VAD bypasses both bottlenecks by directly probing visual evidence, retaining likelihood-level preferences, and structuring them with ordered severity thresholds to derive continuous anomaly rankings.
}
    \label{fig:motivation}
\end{figure*}

As illustrated in Fig.~\ref{fig:motivation}, information useful for anomaly
ranking can be progressively lost as the visual understanding of a frozen VLM
is converted into a scalar score. Caption-mediated pipelines may first
introduce a \emph{representation bottleneck} by compressing visual observations
into text, making cues omitted from the description inaccessible to the
downstream scorer. More importantly, even when the scorer directly observes the video, the \emph{decoding bottleneck} remains. Autoregressive numerical generation
retains only the selected output while discarding the model's relative
preferences over alternative responses. Thus, direct visual evidence alone is insufficient; retaining pre-decoding preferences is also necessary for finer anomaly ranking. Yet likelihood retention alone remains inadequate because anomaly severity is inherently ordered. Let \(S\in[0,1]\) denote a conceptual anomaly-severity variable, where larger values indicate stronger anomaly evidence. Here, ``order'' refers to severity rather than temporal frame order: evidence supporting a severity level of at least 0.8 should also support the less demanding conditions of at least 0.5 and 0.3. Probe-VAD therefore progressively refines the score-extraction interface by retaining direct visual evidence, preserving pre-decoding likelihood preferences, and organizing these preferences through cumulative severity-threshold propositions.

In this paper, we propose \textbf{Probe-VAD}, a training-free ordinal likelihood probing framework that progressively preserves and structures the information required for anomaly ranking. Given a video clip, Probe-VAD directly conditions a frozen VLM on the visual input, avoiding intermediate textual compression. Rather than retaining only a decoded numerical response, it extracts pre-decoding \texttt{YES}/\texttt{NO} likelihood preferences from a set of ordinal queries, each associated with a different severity threshold. These preferences are organized into a cumulative ordinal profile, where higher thresholds represent increasingly stringent anomaly conditions. The resulting profile is then integrated into a continuous anomaly score for fine-grained temporal ranking. Experiments on three mainstream VAD datasets, i.e., UCF-Crime~\cite{sultani2018real}, MSAD~\cite{zhu2024advancing} and XD-Violence~\cite{wu2020not}, validate each design choice: direct visual conditioning outperforms caption conditioning, likelihood-based scoring substantially improves over direct numerical generation, and ordinal modeling provides further gains over flat likelihood scoring. Probe-VAD achieves superior performance on three mainstream benchmarks, while also reducing end-to-end wall-clock time by 59.0\% and artifact size by 87.3\% over the complete caption-based pipeline.

Our contributions are summarized as follows:
\begin{itemize}
    \item We identify the score-extraction interface as an underexplored problem in training-free VLM-based VAD, revealing two key bottlenecks: caption mediation may discard anomaly-relevant visual evidence, while autoregressive numerical decoding can lose fine-grained visual information.

    \item We propose \textbf{Probe-VAD}, a training-free ordinal likelihood probing framework that directly accesses visual evidence, retains likelihood-level model preferences before decoding, and organizes them through ordered severity thresholds to produce continuous anomaly rankings.

    \item Extensive experiments on three VAD benchmarks validate the proposed design principles and demonstrate superior detection performance together with improved computational and storage efficiency.
\end{itemize}

\section{Related Work}
\label{sec:related}

\subsection{Video Anomaly Detection}

Video anomaly detection (VAD)~\cite{wu2026deep} has been extensively studied under one class~\cite{cao2024context,leng2022anomaly} and weakly supervised settings~\cite{zhang2026reconstructive, wu2021learning,wu2024vadclip,zhang2026contextual}. Reconstruction- and prediction-based methods~\cite{hasan2016learning, Liu_2018_CVPR} detect deviations from normal patterns, while weakly supervised approaches~\cite{sultani2018real,tian2021weakly} learn anomaly discrimination from video-level labels. Open-vocabulary VAD extends detection beyond predefined anomaly categories but still requires task-specific optimization~\cite{wu2024open}. RVT~\cite{zhang2026reconstructive} introduces a vision-centric reconstructive objective to supervise visual outputs under a weak-supervised paradigm. More recently, TD-VAD~\cite{zhang2026td} reduces reliance on visual training data by learning from LLM-generated textual sequences. Despite these advances, existing paradigms still rely on task-specific learning from either visual or surrogate supervision, which may limit generalization beyond the training distribution. In contrast, Probe-VAD eliminates task-specific training and directly converts the visually conditioned preferences of a frozen VLM into continuous anomaly-ranking signals.

\subsection{Training-Free Video Anomaly Detection}

Training-free VAD has recently emerged as a promising alternative to task-specific learning. Caption-mediated methods, such as LAVAD~\cite{zanella2024harnessing} and URF-HVAA~\cite{lin2025unified}, translate visual observations into textual descriptions and leverage language models for anomaly reasoning. While effective, anomaly-relevant visual cues omitted during caption generation are no longer accessible to the downstream scorer. Other approaches focus on temporal organization: EventVAD~\cite{shao2025eventvad} performs event-aware segmentation and hierarchical reasoning, while VADTree~\cite{li2025vadtree} organizes long videos into a hierarchical multi-granularity structure. These methods primarily improve how visual evidence is temporally organized rather than how anomaly scores are extracted from a fixed visual clip.
More recent approaches explore direct VLM reasoning and efficient inference. CoReVAD~\cite{lim2026corevad} directly generates segment-level anomaly responses and refines them with temporal context, while Flashback~\cite{lee2025flashback} employs pseudo-scene memory and retrieval to reduce repeated model inference. Related multimodal approaches such as VERA~\cite{ye2025vera} and VADOR~\cite{ozturk2023vador} still involve task-specific learning and therefore fall outside the strict training-free setting considered here.

In contrast, {Probe-VAD} focuses on the \emph{score-extraction interface} of frozen VLMs, directly probing ordered severity propositions and retaining likelihood-level preferences to produce continuous anomaly rankings.

\subsection{Ordinal Modeling and Likelihood-Based Scoring}

Ordinal modeling captures the inherent ordering among discrete labels through
ordered or cumulative predictions. In VAD, Pang \emph{et al.}~
\cite{pang2020selftrained} model anomaly severity using ordinal regression,
while isotonic regression provides a non-parametric mechanism for enforcing
monotonic consistency. Likelihood-based scoring instead exploits model
preferences over candidate responses before decoding, although such
preferences can be sensitive to prompt and label priors~\cite{zhao2021calibrate}.
LogicQA~\cite{kwon2025logicqa} applies a related strategy to image anomaly
detection using constrained VLM questions and answer-token probabilities.

Different from these works, {Probe-VAD} combines ordinal structure
with likelihood probing for training-free VAD. It queries a frozen VLM with
ordered severity thresholds, retains \texttt{YES}/\texttt{NO} continuation
likelihoods, and integrates the resulting cumulative ordinal evidence into
continuous anomaly rankings. This avoids learning an ordinal prediction head
while preserving fine-grained model preferences before decoding.

\section{Methodology}
\label{sec:method}

\subsection{Problem Formulation}
\label{subsec:problem}

Let \(V=\{I_f\}_{f=0}^{F-1}\) denote an untrimmed video of \(F\) frames,
where \(I_f\) is the \(f\)-th frame. The frame-level ground truth is
\(Y=\{y_f\}_{f=0}^{F-1}\), where \(y_f\in\{0,1\}\) indicates whether
frame \(f\) belongs to an annotated anomalous interval. Given a frozen VLM, the goal is to produce a continuous frame-level anomaly score sequence \(\hat{Y}=\{\hat{y}_f\}_{f=0}^{F-1}\), with \(\hat{y}_f\in[0,1]\), where higher scores indicate stronger anomaly evidence. No target-domain optimization, parameter updates, prompt learning, or label-based calibration are involved; ground-truth annotations are used solely for evaluation.

The key challenge is therefore to translate the visual understanding of the frozen VLM into continuous anomaly-ranking signals. Instead of compressing visual evidence into intermediate captions or relying on decoded numerical responses, Probe-VAD directly evaluates a sequence of ordered severity propositions conditioned on the visual clip and extracts their likelihood-level evidence for anomaly scoring.

\subsection{Framework Overview and Clip Construction}
\label{subsec:overview}

\begin{figure*}[!t]
	\centering
	\includegraphics[width=\linewidth]{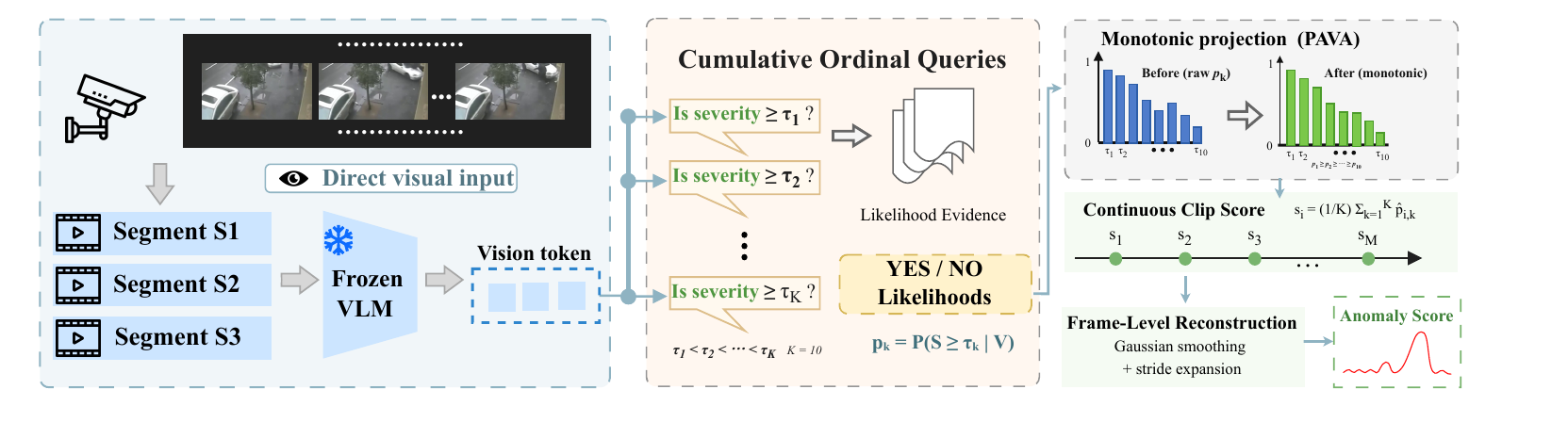}
    \caption{Overview of Probe-VAD. Given video clips, a frozen VLM encodes visual evidence without captions and is probed with ordered severity thresholds. The \texttt{YES}/\texttt{NO} continuation likelihoods form a cumulative ordinal profile, which is integrated into continuous clip-level anomaly scores with the pool-adjacent-violators algorithm (PAVA) enforcing ordinal consistency. The clip-level scores are reconstructed into frame-level anomaly predictions.}
    \label{fig:framework}
\end{figure*}

Fig.~\ref{fig:framework} presents the overall framework of Probe-VAD.
Given an untrimmed video, Probe-VAD consists of four stages:
(1) constructing overlapping temporal clips;
(2) directly encoding each clip with a frozen VideoLLaMA3-7B~\cite{zhang2025videollama};
(3) extracting cumulative ordinal evidence from constrained
\texttt{YES}/\texttt{NO} continuation likelihoods; and
(4) reconstructing the resulting clip-level scores into frame-level anomaly
predictions. The core of Probe-VAD lies in Stages (2) and (3): direct visual
conditioning preserves access to anomaly-relevant visual evidence, while
ordinal likelihood probing converts the VLM's visually conditioned
preferences into continuous anomaly scores. Stages (1) and (4) provide the
temporal interface for processing long untrimmed videos.

Formally, let \(r\) denote the original video frame rate, \(\Delta\) the spacing
in frames between adjacent clip centers, and \(c_i\) the center-frame index
of the \(i\)-th clip. The total number of clips is denoted by \(M\):
\begin{equation}
    c_i=i\Delta,
    \qquad
    i=0,\ldots,M-1,
    \qquad
    M=\left\lfloor\frac{F-1}{\Delta}\right\rfloor+1.
    \label{eq:center_frame}
\end{equation}

Given a temporal window of duration \(L\) seconds, let \(a_i\) and \(b_i\)
denote the start and end times, respectively, for clip \(i\):
\begin{equation}
    a_i=\max\left(0,\frac{c_i}{r}-\frac{L}{2}\right),
    \qquad
    b_i=\min\left(\frac{F}{r},\frac{c_i}{r}+\frac{L}{2}\right).
    \label{eq:clip_window}
\end{equation}

Let \(r_s\) denote the target frame-sampling rate and \(N\) the maximum number of retained RGB frames. The resulting visual input is denoted by \(X_i=\{I_{i,1},\ldots,I_{i,n_i}\}\), where \(I_{i,j}\) denotes the \(j\)-th sampled frame of clip \(i\), and \(n_i\le N\) is the actual number of sampled frames in clip \(i\).

\subsection{Cumulative Ordinal Visual Scoring}
\label{subsec:ordinal}

For the \(i\)-th clip \(X_i\), let \(S_i\in[0,1]\) denote a conceptual latent severity variable representing anomaly evidence, where larger values indicate stronger anomaly evidence. Importantly, \(S_i\) is neither observed nor learned from severity annotations; it serves only to define an ordinal scoring formulation. Rather than asking the frozen VLM to directly decode a single numerical severity value, we characterize \(S_i\) through \(K\) nested threshold propositions, where \(K\) denotes the number of severity thresholds. Let \(\mathcal{T}=\{\tau_k\}_{k=1}^{K}\) denote the ordered threshold grid, where \(\tau_k\) is the \(k\)-th severity threshold:
\begin{equation}
    \tau_k=\frac{k}{K},
    \label{eq:threshold_grid}
\end{equation}
We use \(K=10\), yielding
\(\mathcal{T}=\{0.1,0.2,\ldots,1.0\}\).

For each threshold \(\tau_k\), the frozen VLM evaluates whether the anomaly severity in \(X_i\) reaches or exceeds \(\tau_k\) using a constrained binary query with \texttt{YES}/\texttt{NO} continuations. The query semantics are fixed across thresholds and datasets by a common severity scale, while only the numerical threshold \(\tau_k\) varies. This yields a sequence of ordered severity judgments for extracting likelihood-level ordinal preferences.

For the \(k\)-th threshold, let \(Q_k\) denote the corresponding binary query. Let \(C\in\{\mathrm{YES},\mathrm{NO}\}\) denote a candidate continuation, represented by the token sequence \(c_{1:m_C}\), where \(c_t\) is the token at position \(t\) and \(m_C\) is the number of tokens in candidate \(C\). We denote the frozen VLM by \(\Phi\), with \(P_{\Phi}\) representing its conditional token probability for each candidate continuation token. The conditional log-likelihood of candidate \(C\) for clip \(X_i\) under query \(Q_k\) is denoted by \(\ell^{C}_{i,k}\):
\begin{equation}
    \ell^{C}_{i,k}
    =
    \sum_{t=1}^{m_C}
    \log P_{\Phi}
    \left(
        c_t
        \mid
        X_i,Q_k,c_{<t}
    \right),
    \label{eq:candidate_likelihood}
\end{equation}
where \(c_{<t}\) denotes the candidate tokens preceding position \(t\).
Let \(T>0\) denote the likelihood temperature. We denote by \(p_{i,k}\)
the normalized \texttt{YES} preference for clip \(X_i\) at threshold
\(\tau_k\):
\begin{equation}
    p_{i,k}
    =
    \frac{\exp\left(\ell^{\mathrm{YES}}_{i,k}/T\right)}
    {\exp\left(\ell^{\mathrm{YES}}_{i,k}/T\right)+
     \exp\left(\ell^{\mathrm{NO}}_{i,k}/T\right)}.
    \label{eq:tail_probability}
\end{equation}
The resulting \(p_{i,k}\) measures the VLM's relative support for the proposition that the anomaly severity in \(X_i\) reaches or exceeds \(\tau_k\). We denote the corresponding VLM-induced \emph{tail-evidence proxy} by \(\widehat{P}_{\Phi}(S_i\ge\tau_k\mid X_i)\), defined as
\begin{equation}
    \widehat{P}_{\Phi}(S_i\ge\tau_k\mid X_i)
    \mathrel{:=}p_{i,k}.
    \label{eq:tail_interpretation}
\end{equation}
This notation reflects the cumulative-threshold interpretation rather than a calibrated probability.

This likelihood-based score extraction differs fundamentally from direct
autoregressive generation. A decoded numerical response retains only the
selected output, whereas Eq.~\eqref{eq:tail_probability} preserves the
VLM's relative preference between the constrained continuations. Moreover, the threshold propositions are inherently nested: evidence supporting a higher severity threshold should remain compatible with all lower thresholds. Accordingly, a valid cumulative profile should satisfy
\begin{equation}
    p_{i,1}\ge p_{i,2}\ge\cdots\ge p_{i,K}.
    \label{eq:ordinal_constraint}
\end{equation}
This monotonic structure explicitly captures the ordinal nature of anomaly
severity, rather than treating different severity levels as independent
categories.

\subsection{Monotonic Projection and Cumulative Tail Integration}
\label{subsec:projection}

Although the threshold propositions are ordinally related, their likelihood preferences are not explicitly constrained to satisfy the expected monotonic relation. Let \(\mathbf{p}_i=[p_{i,1},\ldots,p_{i,K}]\) denote the raw threshold-wise severity profile of clip \(X_i\). Because this profile may contain local monotonicity violations, we denote its monotonic projection by \(\hat{\mathbf{p}}_i\), use \(\mathbf{z}=[z_1,\ldots,z_K]\in\mathbb{R}^{K}\) as the optimization variable, and let \(\|\cdot\|_2\) denote the Euclidean norm:
\begin{equation}
    \hat{\mathbf{p}}_i
    =
    \arg\min_{\mathbf{z}\in\mathbb{R}^{K}}
    \|\mathbf{z}-\mathbf{p}_i\|_2^2
    \quad
    \mathrm{s.t.}
    \quad
    1\ge z_1\ge\cdots\ge z_K\ge0.
    \label{eq:pava_projection}
\end{equation}
We solve Eq.~\eqref{eq:pava_projection} using the pool-adjacent-violators algorithm (PAVA), which introduces no learnable parameters. This projection converts the raw threshold-wise preferences into a monotonic cumulative profile consistent with the ordinal structure of anomaly severity. Under the uniform threshold grid and equal-weight aggregation used below, PAVA preserves the final scalar score used in the final ranking; its role is therefore to enforce structural consistency on the intermediate profile rather than alter the ranking statistic.

Let \(\hat{p}_{i,k}\) denote the \(k\)-th component of the projected profile \(\hat{\mathbf{p}}_i\). To convert the cumulative profile into a continuous anomaly score, we draw on the tail-integral identity. For a generic bounded random variable \(S\in[0,1]\) and a continuous severity threshold \(\tau\in[0,1]\), the tail-integral identity gives
\begin{equation}
    \mathbb{E}[S]
    =
    \int_0^1 P(S\ge\tau)\,d\tau.
    \label{eq:tail_integral}
\end{equation}
Motivated by this relation, let \(s_i\) denote the continuous anomaly score of clip \(X_i\), and define \(\tau_0=0\). We aggregate the projected VLM-induced tail evidence across the ordered threshold grid using a right Riemann sum:
\begin{equation}
    s_i
    =
    \sum_{k=1}^{K}
    (\tau_k-\tau_{k-1})\hat{p}_{i,k},
    \qquad
    \tau_0=0.
    \label{eq:continuous_score_general}
\end{equation}
For the uniform grid defined in Eq.~\eqref{eq:threshold_grid},
\(\tau_k-\tau_{k-1}=1/K\), so the continuous anomaly score reduces to
the mean of the projected tail evidence:
\begin{equation}
    s_i
    =
    \frac{1}{K}\sum_{k=1}^{K}\hat{p}_{i,k}.
    \label{eq:continuous_score}
\end{equation}
The resulting \(s_i\in[0,1]\) serves as a continuous ranking statistic that summarizes the VLM's support across increasing anomaly-severity thresholds. The tail-integral identity motivates this aggregation, but neither \(\hat{p}_{i,k}\) nor \(s_i\) is assumed to be probabilistically calibrated.

An important property of the equal-weight formulation is that PAVA preserves the mean of the threshold profile. 
\begin{equation}
    \frac{1}{K}\sum_{k=1}^{K}\hat{p}_{i,k}
    =
    \frac{1}{K}\sum_{k=1}^{K}p_{i,k}.
    \label{eq:pava_sum_invariance}
\end{equation}
This follows directly from the pooling operation of PAVA. For each contiguous block \(\mathcal{B}\) that violates the monotonic constraint, with \(|\mathcal{B}|\) denoting the number of elements in the block, PAVA replaces all entries by their block mean,
\begin{equation}
    \hat{p}_{i,k}
    =
    \frac{1}{|\mathcal{B}|}
    \sum_{j\in\mathcal{B}}p_{i,j},
    \qquad k\in\mathcal{B},
\end{equation}
which preserves the block sum:
\begin{equation}
    \sum_{k\in\mathcal{B}}\hat{p}_{i,k}
    =
    \sum_{k\in\mathcal{B}}p_{i,k}.
\end{equation}
Since the pooled blocks form a disjoint partition of threshold indices, the total sum is preserved. Consequently, under the uniform threshold grid and equal-weight integration in Probe-VAD, PAVA enforces a monotonic, ordinally consistent profile without altering the final scalar anomaly score.

\begin{figure*}[!t]
    \centering
    \includegraphics[width=0.94\textwidth]{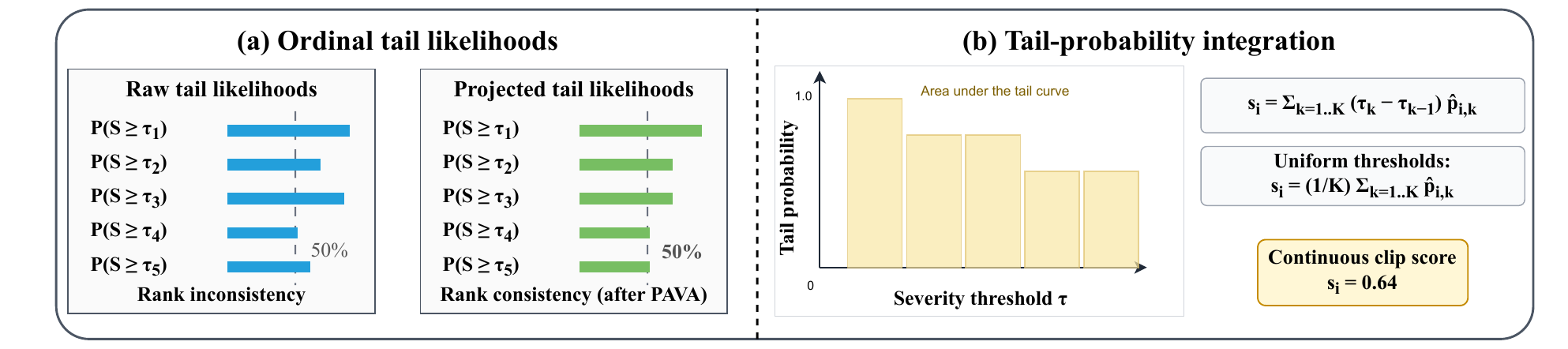}
    \caption{Illustration of ordinal consistency projection and cumulative tail integration. (a) Independently estimated tail likelihoods $P(S\geq\tau_k)$ may violate the required non-increasing order across increasing severity thresholds. PAVA projects them onto a rank-consistent monotone sequence. (b) The projected tail-evidence proxies $\hat{p}_{i,k}$ are integrated over the severity thresholds to obtain a continuous clip-level anomaly score. For uniformly spaced thresholds, the Riemann sum reduces to the mean of the projected tail evidence.}
    \label{fig:tail_integration}
\end{figure*}

Fig.~\ref{fig:tail_integration} illustrates the complementary roles of scalar score and intermediate ordinal profile. The integrated tail evidence determines the clip-level ranking score, while the projected profile provides a threshold-wise representation consistent with cumulative severity semantics.

\subsection{Frame-Level Reconstruction and Efficient Inference}
\label{subsec:frame_inference}

The clip-level scores form a regularly sampled temporal sequence \(\mathbf{s}=[s_0,\ldots,s_{M-1}]\). To obtain frame-level predictions, let \(G_{\sigma}\) denote a unit-sum Gaussian kernel with smoothing width \(\sigma\), let \(*\) denote one-dimensional convolution, and let \(\bar{\mathbf{s}}\) denote the smoothed score sequence. We then apply Gaussian smoothing to suppress local temporal fluctuations:
\begin{equation}
    \bar{\mathbf{s}}=\mathbf{s}*G_{\sigma},
    \label{eq:gaussian_smoothing}
\end{equation}
where \(\sigma\) is measured in units of the clip-score grid. Each smoothed clip score is then assigned to its corresponding \(\Delta\)-frame interval for subsequent frame-level score reconstruction:
\begin{equation}
    \hat{y}_f
    =
    \bar{s}_{\min\left(\left\lfloor f/\Delta\right\rfloor,M-1\right)},
    \qquad
    f=0,\ldots,F-1.
    \label{eq:frame_mapping}
\end{equation}
The sequence is finally cropped to the annotated video length.

Since all \(K\) ordinal queries for a clip share the same visual input,
Probe-VAD encodes each clip only once and reuses the resulting visual
representation across all severity thresholds. Only the threshold-dependent
textual queries vary, allowing compatible queries to be evaluated in batches
and shared prefixes to be cached. These optimizations reduce redundant
computation without altering the scoring formulation in
Eqs.~\eqref{eq:candidate_likelihood}--\eqref{eq:continuous_score}; the
threshold batch size therefore affects computational efficiency and memory
usage, but not the resulting anomaly scores.

\section{Experiments}
\label{sec:experiments}

We evaluate Probe-VAD on three VAD benchmarks through comparisons
with representative methods, controlled analyses of its core scoring
mechanisms, and robustness studies under different visual, linguistic,
and temporal configurations.

\subsection{Experimental Setup}
\label{subsec:exp_setup}

\noindent\textbf{Datasets}.
Experiments are conducted on three widely used video anomaly detection
benchmarks: UCF-Crime~\cite{sultani2018real},
MSAD~\cite{zhu2024advancing}, and
XD-Violence~\cite{wu2020not}. UCF-Crime contains 290 test videos covering 13 real-world anomaly categories. The evaluated MSAD test split consists of 360 videos from diverse indoor and outdoor surveillance scenarios. XD-Violence contains 800 test videos collected from surveillance footage, movies, and online videos, covering six categories of events.

\noindent\textbf{Evaluation Metrics and Protocol}. Frame-level area under the receiver operating characteristic curve (AUC) is adopted as the primary evaluation metric. Following the evaluation convention of prior work~\cite{lin2025unified}, we also adopt frame-level Average Precision (AP) for MSAD and XD-Violence. Ground-truth annotations are used exclusively for performance evaluation and are not involved in prompt construction, candidate specification, temporal sampling, or score calibration. All VLM parameters remain frozen throughout evaluation, and the same default configuration is used across datasets unless otherwise specified, ensuring consistent evaluation across all reported benchmark comparisons.

\noindent\textbf{Implementation Details}.
VideoLLaMA3-7B~\cite{zhang2025videollama} is used as the default frozen
VLM. Unless otherwise specified, each temporal clip spans 10 seconds and
contains at most 10 uniformly sampled frames. Probe-VAD uses ten severity thresholds uniformly distributed over \([0.1,1.0]\), with likelihood temperature \(T=1.0\). Gaussian smoothing with \(\sigma=10\) is used for frame-level score reconstruction. Qwen3-VL-8B-Instruct~\cite{bai2025qwen3vl} is additionally evaluated to assess portability across frozen VLM backbones. The main VLM inference experiments are conducted on NVIDIA A100 GPUs. The severity-structure analyses operate on retained threshold-wise outputs and require no additional VLM inference.

\begin{figure}[!t]
    \centering
    \includegraphics[width=0.98\linewidth]{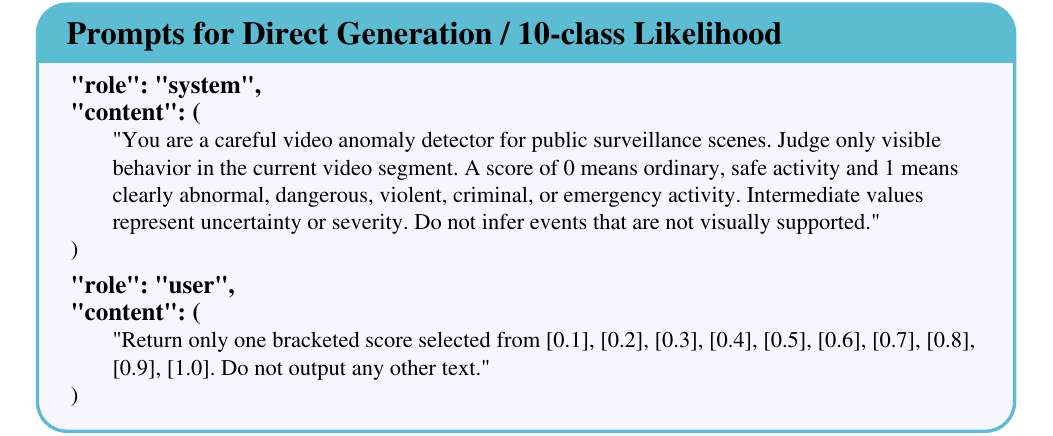}
    \caption{Prompt configuration shared by Direct Generation and
    10-class Likelihood.}
    \label{fig:prompt_numeric}
\end{figure}

\begin{figure}[!t]
    \centering
    \includegraphics[width=0.98\linewidth]{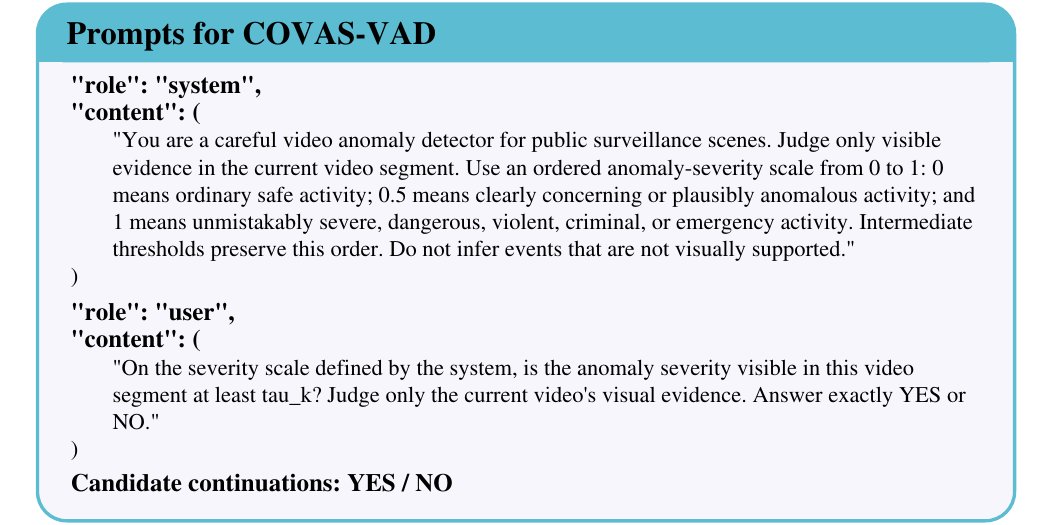}
    \caption{Prompt configuration used by Probe-VAD.}
    \label{fig:prompt_covas}
\end{figure}

\begin{figure}[!t]
    \centering
    \includegraphics[width=0.98\linewidth]{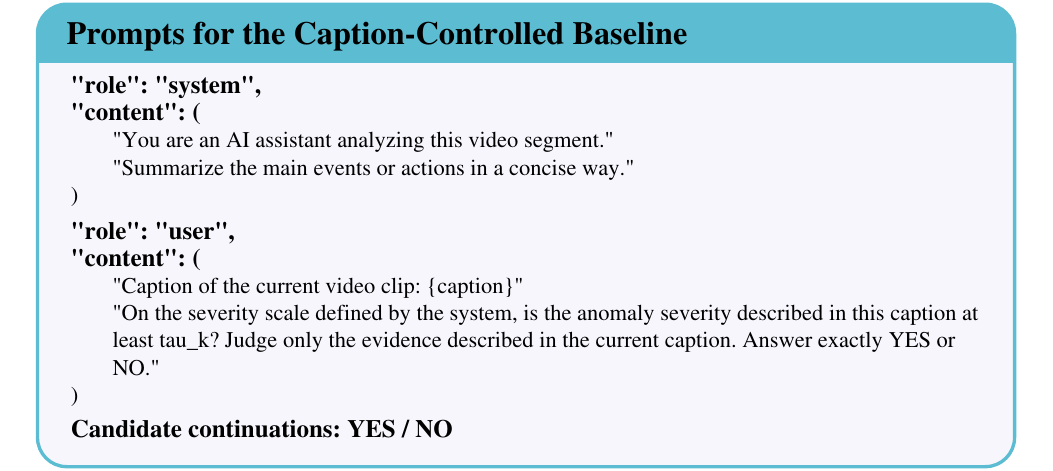}
    \caption{Prompt configuration used by the caption-controlled baseline.}
    \label{fig:prompt_caption}
\end{figure}

\subsection{Comparison with Representative Methods}
\label{subsec:main_results}

Table~\ref{tab:main_results} compares Probe-VAD with representative
training-based and training-free VAD methods on UCF-Crime, XD-Violence,
and MSAD. Results are reported under the evaluation conventions adopted
by the corresponding benchmarks. Since the compared methods differ in
supervision, backbone architecture, temporal processing, and preprocessing,
the table is intended to provide a system-level comparison rather than a
strictly controlled component-wise evaluation.

\begin{table*}[!tbp]
\centering
\caption{\textbf{Comparison with representative VAD methods on UCF-Crime,
XD-Violence, and MSAD.}
\cmark~/~\xmark denote whether a method satisfies the zero-shot and
training-free settings for VAD-specific model parameters.
Results are reported in percentages, and ``--'' denotes unavailable or
non-comparable results. Bold and underlined values indicate the best and
second-best performance within each comparison block, respectively.}
\label{tab:main_results}

\scriptsize
\setlength{\tabcolsep}{3.2pt}
\renewcommand{\arraystretch}{1.03}

\begin{tabular}{@{}l|cc|c|cc|cc@{}}
\toprule

\multirow{2}{*}{\textbf{Method}}
& \multirow{2}{*}{\textbf{Zero-shot}}
& \multirow{2}{*}{\textbf{Training-free}}
& \multicolumn{1}{c|}{\textbf{UCF-Crime}}
& \multicolumn{2}{c|}{\textbf{XD-Violence}}
& \multicolumn{2}{c}{\textbf{MSAD}}
\\

\cmidrule(lr){4-4}
\cmidrule(lr){5-6}
\cmidrule(l){7-8}

&
&
&
\textbf{AUC (\%)}
& \textbf{AUC (\%)}
& \textbf{AP (\%)}
& \textbf{AUC (\%)}
& \textbf{AP (\%)}
\\
\midrule

Sultani et al.~\cite{sultani2018real}
& \xmark & \xmark
& 77.92 & -- & 73.20 & -- & -- \\

GODS~\cite{wang2019gods}
& \xmark & \xmark
& 70.46 & 61.56 & -- & -- & -- \\

RTFM~\cite{tian2021weakly}
& \xmark & \xmark
& 83.31 & -- & 77.81
& \underline{86.70} & \textbf{66.30} \\

AccI-VAD~\cite{reiss2022attribute}
& \xmark & \xmark
& -- & -- & -- & -- & -- \\

CLIP-TSA~\cite{joo2023clip}
& \xmark & \xmark
& 87.58 & -- & {82.19}
& -- & -- \\

MGFN~\cite{chen2023mgfn}
& \xmark & \xmark
& 86.98 & -- & 80.11
& 85.00 & 63.50 \\

STPrompt~\cite{wu2024weakly}
& \xmark & \xmark
& {88.08} & -- & --
& -- & -- \\

OVVAD~\cite{wu2024open}
& \xmark & \xmark
& 86.40 & -- & 66.53
& -- & -- \\

Holmes-VAU~\cite{holmesvau2025}
& \xmark & \xmark
& \textbf{88.96} & -- & \textbf{87.68}
& -- & -- \\

MULDE~\cite{micorek2024mulde}
& \xmark & \xmark
& 78.50 & -- & --
& -- & -- \\

EGO~\cite{ding2025learnable}
& \xmark & \xmark
& 81.71 & -- & 65.77
& \textbf{87.30} & \underline{64.40} \\


RVT~\cite{zhang2026reconstructive}
& \xmark & \xmark
& \underline{88.13}
& --
& \underline{85.77}
& -- & -- \\

VERA~\cite{ye2025vera}
& \xmark & \cmark
& 86.55 & \underline{88.26} & 70.54
& -- & -- \\


TD-VAD~\cite{zhang2026td}
& \cmark & \xmark
& 80.82
& \textbf{89.50}
& 75.83
& -- & -- \\

\midrule


UR-DMU (ZS)~\cite{zhou2023dual}
& \cmark & \cmark
& -- & -- & --
& 74.30 & 53.40 \\

CLIP (ZS)~\cite{radford2021learning}
& \cmark & \cmark
& 53.16 & 38.21 & 17.83
& -- & -- \\

LLaVA-1.5 (ZS)~\cite{liu2024improved}
& \cmark & \cmark
& 72.84 & 79.62 & 50.26
& -- & -- \\

VideoLLaMA3-7B + Llama3.1-8B (ZS)
~\cite{zhang2025videollama,grattafiori2024llama}
& \cmark & \cmark
& -- & -- & --
& 78.70 & 68.50 \\

GLM-4.1V-9B-Thinking (ZS CoT)\textsuperscript{\(\ddagger\)}
& \cmark & \cmark
& 61.80 & 72.73 & 52.93
& -- & -- \\

LAVAD~\cite{zanella2024harnessing}
& \cmark & \cmark
& 80.28 & 85.36 & 62.01
& -- & -- \\

VADTree\textsuperscript{\S}~\cite{li2025vadtree}
& \cmark & \cmark
& 84.74
& 90.44
& 67.82
& --
& -- \\

PANDA~\cite{yang2026panda}
& \cmark & \cmark
& \underline{84.89}
& --
& \underline{70.16}
& --
& -- \\

URF-HVAA (fixed-constant setting)~\cite{lin2025unified}
& \cmark & \cmark
& 84.28
& \underline{91.34}
& 68.07
& \underline{85.90}
& \underline{76.40} \\

\best{Probe-VAD (Ours)}
& \cmark & \cmark
& \best{86.27}
& \best{92.11}
& \best{75.36}
& \best{87.55}
& \best{78.43} \\

\bottomrule
\end{tabular}

\vspace{1mm}

\begin{minipage}{0.99\textwidth}
\scriptsize

\textsuperscript{\(\ddagger\)}
Zero-shot chain-of-thought (CoT) inference VAD performance using GLM-4.1V-9B-Thinking~\cite{hong2025glm}.

\textsuperscript{\S}
VADTree reports MSAD results on a different 240-video evaluation split (120 normal and 120 anomalous videos), whereas Probe-VAD is evaluated on the 360-video split (120 normal and 240 anomalous videos). Its MSAD values are therefore omitted from the directly comparable ranking.

\end{minipage}

\end{table*}

Among the zero-shot and training-free methods included in
Table~\ref{tab:main_results}, Probe-VAD achieves the best performance
across all five directly comparable metrics, reaching 86.27\% AUC on
UCF-Crime, 92.11\% AUC and 75.36\% AP on XD-Violence, and
87.55\% AUC and 78.43\% AP on MSAD. Relative to the strongest
comparable baselines in this setting, the corresponding improvements are
1.38 points on UCF-Crime, 0.77 AUC and 5.20 AP points on
XD-Violence, and 1.65 AUC and 2.03 AP points on MSAD. The consistent direction of improvement across the three benchmarks suggests that the effectiveness of Probe-VAD is not restricted to a single anomaly distribution. The magnitude of the gain, however, varies across
datasets, which is expected because the three benchmarks differ substantially in scene composition, anomaly duration, event complexity, and class imbalance. Probe-VAD does not explicitly model these dataset-specific properties; instead, its advantage arises from changing how visually conditioned evidence is exposed to the anomaly scorer. This makes the consistent improvement across heterogeneous benchmarks particularly relevant to the proposed score-extraction perspective.

Compared with caption-mediated pipelines, Probe-VAD preserves the original visual evidence until the scoring stage, while compared with direct generative approaches it retains response preferences before a single output is selected. Both differences increase the amount of information available for ranking temporal segments. The main results are therefore consistent with our central hypothesis that the performance of a frozen VLM is determined not only by what visual information it encodes, but also by how this information is converted into an anomaly score. The improvement is  pronounced in XD-Violence AP. AUC measures pairwise ordering between positive and negative samples over the full score range, whereas precision--recall evaluation is more sensitive to the quality of positive-sample ranking under class imbalance. The larger AP gain therefore suggests that Probe-VAD improves not only the global ordering of frames but also the concentration of anomalous frames toward the high-score region. This behavior is consistent with the higher score resolution obtained
by likelihood-based inference, which reduces the large tied groups induced by discrete numerical generation.
Despite using no VAD-specific parameter optimization, Probe-VAD remains
competitive with several training-based approaches. This observation
suggests that useful anomaly information is already accessible from frozen VLMs, while the mechanism used to convert this information into continuous scores strongly affects how effectively it can be exploited. Since the methods in Table~\ref{tab:main_results} differ in backbone, temporal processing, and preprocessing protocols, this comparison should be interpreted as a system-level evaluation. The following controlled analyses isolate the contributions of the proposed scoring design more directly.

\subsection{Ablation Studies}
\label{sec:ablation}

Unless otherwise specified, all controlled comparisons use the same frozen VLM, temporal sampling, and frame-level reconstruction protocol, with only the factor under investigation being changed. We organize the analysis around three questions: (1) whether retaining direct visual conditioning improves over caption-mediated evidence, (2) whether likelihood-level preferences provide a more informative scoring interface than decoded outputs, and (3) how the resulting ordinal representation behaves under changes in severity discretization, linguistic formulation, visual context, and post-processing.

\noindent\textbf{Effect of Visual Conditioning}.
We compare direct visual conditioning with a caption-conditioned variant
under the same ordinal likelihood scoring protocol. As shown in
Table~\ref{tab:controlled_caption}, direct visual conditioning improves
AUC by 5.57, 0.76, and 0.41 points on UCF-Crime, MSAD, and
XD-Violence, respectively, supporting the representation-bottleneck
hypothesis in Sec.~\ref{sec:introduction}. Caption generation compresses rich visual observations into compact textual descriptions. While this may preserve dominant event semantics, weaker ranking-relevant cues can be omitted or coarsely expressed. Once visually distinct clips are mapped to similar captions, the downstream scorer cannot recover these lost distinctions. Direct visual conditioning avoids this intermediate compression and therefore preserves a richer basis for continuous anomaly ranking. The larger gain on UCF-Crime further suggests that the impact of caption mediation is dataset dependent. However, the aggregate results do not isolate which specific visual factors cause this difference; we therefore interpret the experiment as evidence that caption mediation can remove ranking-relevant information, rather than attributing the gain to any particular visual cue.
\begin{table}[!t]
    \centering
    \caption{Effect of visual conditioning. Results are frame-level AUC (\%).}
    \label{tab:controlled_caption}
    \footnotesize
    \setlength{\tabcolsep}{7pt}
    \begin{tabular}{@{}ccc@{}}
        \toprule
        Dataset & Input & AUC \\
        \midrule
        UCF-Crime & Caption & 80.70 \\
        & Video & \textbf{86.27} \\
        \addlinespace
        MSAD & Caption & 86.79 \\
        & Video & \textbf{87.55} \\
        \addlinespace
        XD-Violence & Caption & 91.70 \\
        & Video & \textbf{92.11} \\
        \bottomrule
    \end{tabular}
\end{table}

\noindent\textbf{Effect of the Scoring Interface}.
To isolate the effect of score extraction from visual understanding,
Direct Generation and 10-class Likelihood use identical visual inputs,
numerical-severity prompts, and candidate spaces. Direct Generation retains the decoded numerical response, whereas 10-class Likelihood preserves the normalized likelihood distribution over the same ten numerical severity candidates and uses their expected severity as the anomaly score. As shown in Table~\ref{tab:scoring_interface}, replacing Direct Generation with 10-class Likelihood improves AUC by 14.23, 8.48, and 5.59 points on UCF-Crime, MSAD, and XD-Violence, respectively. Because the two variants share the same visual evidence, prompt, and numerical candidate space, this large gap primarily reflects whether the VLM's response preferences are retained before autoregressive decoding. Direct Generation reduces the model response to a single selected severity value and therefore discards information about the relative support assigned to alternative candidates. This creates a fundamental mismatch with the evaluation objective of VAD. Frame-level AUC depends primarily on the relative ordering of anomaly scores, whereas a discrete generative interface forces many temporally distinct clips onto the same small set of numerical values. Once two clips receive the same decoded severity, any difference in the VLM's underlying confidence or preference margin becomes invisible to the ranking metric.

This effect is clearly reflected in the empirical prediction distribution. On UCF-Crime, Direct Generation maps all 69,634 evaluated clips to only ten possible scores, and the most frequent value alone accounts for 46.78\% of all predictions. Thus, the effective resolution of the generated score is considerably lower than the nominal ten-level output space. Large tied groups make it impossible to rank many clips that the VLM may internally regard as different. Likelihood-based scoring avoids this collapse by retaining the relative support assigned to all candidate severities before decoding. Two clips that would receive the same generated value can therefore remain distinguishable
through differences in their likelihood distributions. The resulting score acts as a continuous preference statistic rather than a discrete decision, which is substantially better aligned with the ranking-based objective of VAD. The gains of 14.23, 8.48, and 5.59 AUC points therefore provide strong evidence that a large portion of anomaly-relevant information is present in the frozen VLM before decoding but is lost when only the final generated response is retained.
Probe-VAD further improves over 10-class Likelihood by 1.84, 0.78, and
0.34 AUC points on UCF-Crime, MSAD, and XD-Violence, respectively.
The distinction here is more subtle. Flat likelihood scoring treats the ten severity values as competing alternatives, although their semantics are intrinsically ordered. For example, strong support for a high severity value does not explicitly encode that weaker severity conditions should also be satisfied. Probe-VAD instead decomposes severity into nested threshold propositions, converting the prediction problem from competition among independent numerical labels into a sequence of cumulative judgments.

This cumulative formulation better matches the structure of anomaly severity: evidence supporting a high threshold should remain compatible with all lower thresholds. The additional improvement over flat likelihood scoring therefore suggests that preserving model preferences is not sufficient by itself; how those preferences are structured also affects the quality of the resulting ranking signal. The relative magnitudes of the improvements reveal a clear two-stage effect. The dominant gain arises from retaining pre-decoding model preferences,
whereas cumulative ordinal probing provides a smaller but consistent
additional benefit by imposing a severity-aware factorization of those
preferences. Because the candidate semantics and query factorization change jointly between 10-class Likelihood and Probe-VAD, the latter gain should be attributed to the cumulative ordinal interface as a whole rather than to ordinality alone. Moreover, PAVA cannot explain the performance difference because the equal-weight aggregation preserves the final scalar score
exactly.
\begin{table}[H]
    \centering
    \caption{Effect of different anomaly-scoring interfaces. Results are frame-level AUC (\%).}
    \label{tab:scoring_interface}
    \footnotesize
    \setlength{\tabcolsep}{2.6pt}
    \begin{tabular}{@{}cccc@{}}
        \toprule
        Scoring Interface & UCF & MSAD & XD \\
        \midrule
        Direct Generation & 70.20 & 78.29 & 86.18 \\
        10-class Likelihood & 84.43 & 86.77 & 91.77 \\
        Probe-VAD & \textbf{86.27} & \textbf{87.55} & \textbf{92.11} \\
        \bottomrule
    \end{tabular}
\end{table}

\noindent\textbf{Effect of Multi-frame Context and Temporal Order}.
We compare the default ordered multi-frame input with a shuffled-frame
variant containing the same observations and a single-frame variant using only the clip center frame. As shown in Table~\ref{tab:temporal_order}, ordered multi-frame input improves AUC from 83.20\% to 86.27\%, corresponding to a gain of 3.07 percentage points over the single-frame setting. In contrast, shuffling the same frames reduces performance by only 0.13 points. The pronounced gap between single-frame and multi-frame input indicates that the extracted likelihood signal benefits substantially from observing multiple
visual states. A single frame provides only an instantaneous observation and can be ambiguous when normal and abnormal events share similar appearance. Multiple frames expose changes in actors, objects, interactions, and scene configuration, allowing the frozen VLM to accumulate contextual evidence that is unavailable from a single observation.

However, the difference between ordered and shuffled multi-frame input is only 0.13 AUC points. This contrast is informative: most of the benefit appears to originate from the availability of multiple complementary visual states rather than from precise temporal ordering itself. In other words, the current VLM can exploit temporal coverage and cross-frame context, but the experiment provides only limited evidence that it performs strong fine-grained sequence reasoning over those frames. This observation also clarifies the role of Probe-VAD. The proposed method does not attempt to introduce a new temporal encoder; instead, it improves how the evidence already exposed by the frozen VLM is converted into an anomaly score. More sophisticated temporal modeling is therefore complementary rather than competing with ordinal likelihood probing, and may further improve performance for anomalies whose interpretation depends strongly on action order or long-range temporal dependencies.
\begin{table}[!t]
    \centering
    \caption{Effect of multi-frame context and temporal order on UCF-Crime. Results are frame-level AUC (\%).}
    \label{tab:temporal_order}
    \footnotesize
    \setlength{\tabcolsep}{10pt}
    \begin{tabular}{@{}cc@{}}
        \toprule
        Input & AUC \\
        \midrule
        Single-frame & 83.20 \\
        Shuffled frames & 86.14 \\
        Ordered frames & \textbf{86.27} \\
        \bottomrule
    \end{tabular}
\end{table}

\noindent\textbf{Effect of the VLM Backbone}.
We replace VideoLLaMA3-7B with Qwen3-VL-8B-Instruct using the
same cumulative ordinal scoring interface. As shown in Table~\ref{tab:backbone}, the two backbones achieve comparable AUC on
all three datasets. Qwen3-VL slightly improves MSAD AUC from 87.55\% to 87.74\%, while VideoLLaMA3 performs better on UCF-Crime and XD-Violence. These results indicate that ordinal likelihood profiles
can be extracted from multiple frozen VLMs and that the proposed scoring
interface is not tied to a single backbone. The remaining differences also show that Probe-VAD does not eliminate the influence of the underlying representation. The scoring interface determines how visually conditioned preferences are exposed, whereas the quality of
those preferences still depends on the visual encoder, multimodal alignment, and linguistic priors of the frozen VLM. 

This distinction is particularly visible on XD-Violence. The two backbones obtain nearly identical AUC values (92.11\% versus 91.97\%), yet their AP values differ substantially (75.36\% versus 70.69\%). Similar AUC therefore does not imply identical behavior throughout the ranked score distribution. The larger AP difference suggests that the two VLMs differ more strongly in how anomalous frames are concentrated toward the highest score region, even when their overall positive--negative ordering remains similar. We therefore interpret this experiment as evidence for portability of the scoring interface across the evaluated VLMs, while recognizing that absolute performance remains backbone dependent.
\begin{table}[!t]
    \centering
    \caption{Effect of the frozen VLM backbone. Results are percentages.}
    \label{tab:backbone}
    \scriptsize
    \setlength{\tabcolsep}{3pt}
    \begin{tabular}{@{}cccc@{}}
        \toprule
        Backbone & Dataset & AUC & AP \\
        \midrule
        \multirow{3}{*}{VideoLLaMA3-7B}
        & UCF-Crime & \textbf{86.27} & -- \\
        & MSAD & 87.55 & \textbf{78.43} \\
        & XD-Violence & \textbf{92.11} & \textbf{75.36} \\
        \addlinespace
        \multirow{3}{*}{\shortstack[l]{Qwen3-VL-8B}}
        & UCF-Crime & 84.79 & -- \\
        & MSAD & \textbf{87.74} & 77.67 \\
        & XD-Violence & 91.97 & 70.69 \\
        \bottomrule
    \end{tabular}
\end{table}

\noindent\textbf{Effect of Candidate Semantics}.
We evaluate different candidate formulations on MSAD while keeping the
visual input, threshold grid, and scoring procedure unchanged. As shown in
Table~\ref{tab:candidate_ablation}, A/B candidates achieve 86.06\%
AUC, label-swap averaging reaches 86.21\%, polarity-pair averaging
reaches 87.27\%, and the default \texttt{YES}/\texttt{NO} formulation
achieves the highest result of 87.55\%. The relatively limited variation indicates that Probe-VAD is not critically dependent on a specific candidate vocabulary. Nevertheless, the consistent difference between neutral A/B labels and semantically meaningful \texttt{YES}/\texttt{NO} responses indicates that candidate semantics
influence the likelihood signal. A plausible explanation is that
\texttt{YES}/\texttt{NO} directly expresses the truth value of the queried threshold proposition, whereas A/B introduces an arbitrary mapping between token identity and semantic judgment. The latter may expose the likelihood estimate more strongly to token- or position-specific priors.

Label-swap averaging slightly improves the A/B formulation, consistent with reducing arbitrary label preference. Polarity-pair averaging further narrows the gap but requires additional queries and still does not exceed the default formulation. These results support the semantically aligned \texttt{YES}/\texttt{NO} candidate pair as a simple and effective default.

\begin{table}[H]
    \centering
    \caption{Effect of candidate semantics on MSAD.}
    \label{tab:candidate_ablation}
    \footnotesize
    \setlength{\tabcolsep}{8pt}
    \begin{tabular}{@{}cc@{}}
        \toprule
        Candidate formulation & AUC \\
        \midrule
        A/B & 86.06 \\
        A/B + label swap & 86.21 \\
        YES/NO + polarity pair & 87.27 \\
        YES/NO (default) & \textbf{87.55} \\
        \bottomrule
    \end{tabular}
\end{table}

\noindent\textbf{Effect of Prompt Formulation}.
We evaluate several semantically similar threshold-query formulations on MSAD while fixing the visual input, severity scale, candidate responses, and threshold grid. As shown in Table~\ref{tab:prompt_sensitivity}, AUC ranges from 86.95\% to 87.76\%, while AP ranges from 78.28\% to 78.97\%. The narrow performance range indicates that the anomaly ranking extracted by Probe-VAD is relatively stable under moderate linguistic reformulation. At the same time, the non-zero variation confirms that conditional likelihoods are not invariant to wording: semantically similar prompts can  induce different response-preference distributions in the frozen VLM.

The \textit{Rated above} formulation achieves the highest numerical result, but exceeds the pre-specified default by only 0.21 AUC and 0.54 AP points. Importantly, this post-hoc best variant is not used for the main results. Retaining a common default prompt across all benchmarks avoids dataset-specific prompt selection and indicates that the reported performance does not depend on choosing the best test-set paraphrase.

\begin{table}[H]
    \centering
    \caption{Effect of threshold-query wording on MSAD.}
    \label{tab:prompt_sensitivity}
    \footnotesize
    \setlength{\tabcolsep}{4.5pt}
    \begin{tabular}{@{}ccc@{}}
        \toprule
        Prompt variant & AUC & AP \\
        \midrule
        Default & 87.55 & 78.43 \\
        Visible evidence & 87.30 & 78.28 \\
        Reach level & 86.95 & 78.62 \\
        No less than & 86.97 & 78.42 \\
        Rated above & \textbf{87.76} & \textbf{78.97} \\
        \bottomrule
    \end{tabular}
\end{table}

\noindent\textbf{Analysis of the Ordinal Severity Structure}. We further examine the ordinal severity representation to assess its
sensitivity to severity discretization and characterize how the retained
evidence varies across severity regions. Specifically, we analyze three
aspects: severity resolution, threshold placement, and the responses of
individual thresholds and adjacent severity intervals. All analyses are
conducted on MSAD using the same scoring and frame-level reconstruction
protocol as the default configuration.

For the resolution analysis, we vary the number of ordered thresholds using $K\in\{1,2,5,10\}$ while following the predefined rule $\tau_k=k/K$. Consequently, the $K=1$ setting contains only $\tau=1.0$ and does not involve test-set-dependent threshold selection. To assess sensitivity to threshold placement, we further fix $K=5$ and redistribute the thresholds toward different portions of the severity axis.For non-uniform grids, the final score is computed using the interval-width weighted integration in Eq.~\eqref{eq:continuous_score_general}. Beyond the final integrated score, we also characterize the internal severity profile. For two adjacent thresholds, let \(d_{i,k}\) denote the local evidence transition between \(\tau_k\) and \(\tau_{k+1}\), defined as
\begin{equation}
    d_{i,k}
    =
    \hat{p}_{i,k}
    -
    \hat{p}_{i,k+1},
    \label{eq:severity_band}
\end{equation}
where $d_{i,k}\geq0$ because the PAVA-projected profile is non-increasing. Since $\hat{p}_{i,k}$ represents a VLM-induced tail-evidence proxy rather than a calibrated probability, $d_{i,k}$ is interpreted only as an interval-level evidence transition, rather than probability mass. Threshold-wise and interval-wise AUC/AP values are used solely for structural diagnosis and are not used for model or threshold selection.

\begin{table}[H]
    \centering
    \caption{Effect of the number of severity thresholds on MSAD.
    All threshold sets follow $\tau_k=k/K$.}
    \label{tab:threshold_number}
    \footnotesize
    \setlength{\tabcolsep}{4.0pt}
    \begin{tabular}{@{}cccc@{}}
        \toprule
        $K$ & Thresholds & AUC & AP \\
        \midrule
        1  & $\{1.0\}$ & 86.97 & 76.77 \\
        2  & $\{0.5,1.0\}$ & 87.15 & 77.51 \\
        5  & $\{0.2,0.4,\ldots,1.0\}$ & 87.41 & 78.12 \\
        10 & $\{0.1,0.2,\ldots,1.0\}$ & \textbf{87.55}
           & \textbf{78.43} \\
        \bottomrule
    \end{tabular}
\end{table}

Table~\ref{tab:threshold_number} shows that increasing the severity resolution consistently improves both ranking metrics. Relative to the single-threshold configuration, $K=10$ improves AUC from 86.97\% to 87.55\% and AP from 76.77\% to 78.43\%, corresponding to absolute gains of 0.58 and 1.66 percentage points, respectively. This indicates that representing anomaly severity with multiple ordered propositions provides additional information beyond a single severity decision.

Most of the improvement is already obtained with a moderate severity resolution. Specifically, $K=5$ accounts for approximately 76\% of the total AUC gain and 81\% of the total AP gain observed between $K=1$ and $K=10$. Increasing the resolution from $K=5$ to $K=10$ yields a further 0.14 AUC points and 0.31 AP points. These results favor a multi-threshold representation while indicating diminishing returns once the severity axis is sufficiently populated. We retain $K=10$ because it achieves the best overall performance and provides a finer profile for structural analysis.

Table~\ref{tab:severity_grid} further examines whether the performance gain depends on the particular placement of the thresholds. Across the four $K=5$ configurations, AUC varies within 87.28--87.49\% and AP within 77.76--78.15\%. Notably, the high-focused grid attains the highest numerical result, but exceeds the  uniform grid by only 0.08 AUC and 0.03 AP points.

\begin{table}[H]
    \centering
    \caption{\textbf{Sensitivity to severity-threshold placement on MSAD.}
    All configurations use $K=5$.}
    \label{tab:severity_grid}
    \scriptsize
    \setlength{\tabcolsep}{3.2pt}
    \begin{tabular}{@{}cccc@{}}
        \toprule
        Grid & Thresholds & AUC & AP \\
        \midrule
        Uniform
        & $\{0.2,0.4,0.6,0.8,1.0\}$
        & 87.41 & 78.12 \\

        Low-focused
        & $\{0.1,0.2,0.3,0.6,1.0\}$
        & 87.28 & 77.76 \\

        Mid-focused
        & $\{0.1,0.3,0.5,0.7,1.0\}$
        & 87.31 & 77.90 \\

        High-focused
        & $\{0.1,0.4,0.7,0.9,1.0\}$
        & \textbf{87.49} & \textbf{78.15} \\
        \bottomrule
    \end{tabular}
\end{table}

The narrow variation suggests that the observed performance is not narrowly tied to a specific partition of the severity axis. We therefore retain the uniform grid as a simple, stable, and dataset-independent configuration rather than selecting the numerically best alternative on the test set. This choice also avoids introducing additional dataset-specific tuning into the default setting. This experiment should consequently be interpreted as a sensitivity analysis of threshold placement, rather than as a search for an optimal severity grid.

The following severity-structure analyses are performed offline on retained threshold-wise outputs and are used only to characterize the internal ordinal response structure. They require no additional VLM inference and do not affect model selection or the reported benchmark results.

\begin{figure}[!t]
    \centering
    \includegraphics[width=0.98\linewidth]
    {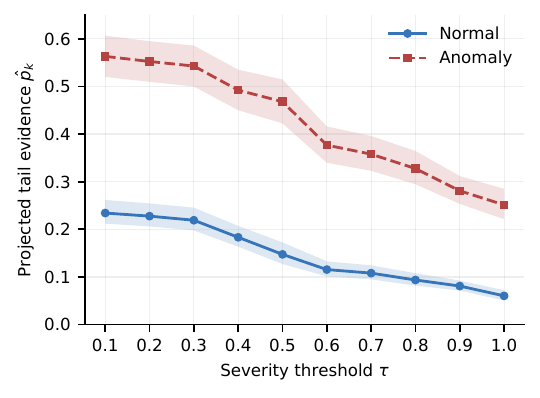}
   \caption{\textbf{Cumulative severity profiles on MSAD.}
Group-mean PAVA-projected tail-evidence proxy \(\hat{p}_{i,k}\) is shown separately for normal and anomalous clips across the ten ordered severity thresholds. Shaded regions denote 95\% confidence intervals obtained by
video-cluster bootstrap over the complete 360-video MSAD test split. Because monotonicity is imposed by PAVA, the informative pattern is the threshold-dependent separation between the two groups rather than the decreasing shape itself.}
    \label{fig:severity_profile}
\end{figure}

Fig.~\ref{fig:severity_profile} visualizes the mean PAVA-projected
threshold-wise responses of normal and anomalous clips. Since the
non-increasing shape is explicitly imposed by isotonic projection,
monotonicity itself is not treated as an empirical finding. Instead, the
relevant observation is the separation between the two groups: anomalous
clips retain higher mean tail evidence across the evaluated severity range,while the magnitude of this separation varies across thresholds.

Threshold-wise diagnostics further show that different summary statistics are maximized at different severity levels. The largest mean
anomaly--normal difference occurs at $\tau=0.1$ (0.3294), whereas the highest standalone AUC is observed at $\tau=0.8$ (87.54\%) and the highest standalone AP at $\tau=0.4$ (78.39\%). Thus, the threshold that maximizes average group separation does not necessarily provide the strongest sample-wise ranking across different evaluation metrics.

This difference reflects the distinct quantities measured by these statistics. Mean separation characterizes the displacement between the two response distributions, whereas AUC depends on pairwise ranking and AP is sensitive to the ordering of positive samples. Their different optima therefore indicate that no single threshold statistic fully characterizes the information contained in the ordinal severity profile.

\begin{table}[!t]
    \centering
    \caption{\textbf{Interval-wise analysis of the ordinal severity structure on MSAD.}
Mean \(d_{i,k}\) denotes the interval evidence transition averaged over all evaluated clips, while \(\Delta_{\mathrm{A-N}}\) denotes the difference between anomalous and normal interval evidence.}
    \label{tab:severity_band}
    \scriptsize
    \setlength{\tabcolsep}{2.7pt}
    \begin{tabular}{@{}ccccc@{}}
        \toprule
        Band &  Mean $d_{i,k}$ & $\Delta_{\mathrm{A-N}}$
             & AUC (\%) & AP (\%) \\
        \midrule
        0.1--0.2 & 0.0080 &  0.0042 & 69.23 & 51.18 \\
        0.2--0.3 & 0.0089 &  0.0010 & 54.18 & 41.96 \\
        0.3--0.4 & 0.0410 &  0.0147 & 68.82 & 52.86 \\
        0.4--0.5 & 0.0318 & -0.0116 & 34.36 & 31.75 \\
        0.5--0.6 & \textbf{0.0529} & \textbf{0.0593}
                  & 83.09 & 64.79 \\
        0.6--0.7 & 0.0115 & 0.0115 & 75.70 & 60.89 \\
        0.7--0.8 & 0.0200 & 0.0158 & 80.34 & 65.40 \\
        0.8--0.9 & 0.0248 & 0.0341
                  & \textbf{83.77} & \textbf{74.79} \\
        0.9--1.0 & 0.0238 & 0.0088 & 55.32 & 53.00 \\
        \bottomrule
    \end{tabular}
\end{table}

The interval-wise results in Table~\ref{tab:severity_band} reveal a pronounced non-uniform structure along the severity axis. The $0.5$--$0.6$ interval exhibits both the largest mean transition magnitude (0.0529) and the largest anomalous-minus-normal difference (0.0593), whereas the $0.8$--$0.9$ interval provides the strongest standalone ranking signal, with 83.77\% AUC and 74.79\% AP. Consistent with the threshold-wise analysis, the interval with the strongest average group separation is therefore not the one with the strongest ranking performance.

The $0.4$--$0.5$ interval provides an informative counterexample. Its $\Delta_{\mathrm{A-N}}$ is $-0.0116$, and its standalone AUC is 34.36\%, showing that this local transition does not independently behave as a positively oriented anomaly-ranking signal. This does not contradict the PAVA monotonicity constraint: $d_{i,k}$ remains non-negative within each
projected profile, whereas $\Delta_{\mathrm{A-N}}$ compares its average magnitude between normal and anomalous groups. The resulting heterogeneity suggests that individual intervals should not be interpreted as interchangeable binary detectors; instead, they describe different local changes within the cumulative severity representation.

Overall, these analyses indicate that Probe-VAD is only mildly sensitive to the evaluated discretization choices, while its internal severity representation exhibits substantial threshold- and interval-dependent variation. This behavior supports treating the ordered responses as a joint cumulative profile rather than selecting a single severity level post hoc.

\noindent\textbf{Effect of Clip-internal Frame Sampling}. We compare uniform sampling with two center-focused strategies on MSAD under the same ten-frame budget. Uniform, Center-4s, and Center-2s achieve 87.55\%, 87.60\%, and 87.54\% AUC, respectively, with a maximum difference of only 0.06 percentage points. The nearly identical results indicate limited sensitivity to the evaluated frame-sampling strategies across different temporal focuses. We retain uniform sampling as the default because it provides more consistent temporal coverage throughout the clip without introducing additional sampling assumptions or dataset-specific preferences.

\begin{table}[H]
    \centering
    \caption{\textbf{Effect of clip-internal frame sampling on MSAD.}
    Each strategy uses ten input frames.}
    \label{tab:center_sampling}
    \footnotesize
    \setlength{\tabcolsep}{8pt}
    \begin{tabular}{@{}lcc@{}}
        \toprule
        Sampling Strategy & AUC (\%) & $\Delta$AUC \\
        \midrule
        Uniform (default) & 87.55 & -- \\
        Center-4s & \textbf{87.60} & +0.05 \\
        Center-2s & 87.54 & -0.01 \\
        \bottomrule
    \end{tabular}
\end{table}

\noindent\textbf{Effect of Temporal Smoothing}. We evaluate Gaussian smoothing widths \(\sigma\in\{0,5,10,20\}\) on MSAD to examine the sensitivity of frame-level score reconstruction. As shown in Table~\ref{tab:smoothing_sensitivity}, the performance remains relatively stable under moderate smoothing strengths. The unsmoothed setting achieves 88.13\% AUC, while \(\sigma=5\) yields 88.18\%, corresponding to only a 0.05-point difference. The default \(\sigma=10\) obtains 87.55\% AUC, whereas excessive smoothing with \(\sigma=20\) reduces performance to 86.65\%.

These results indicate that Probe-VAD is not critically dependent on a narrowly tuned smoothing parameter within a moderate range. Gaussian smoothing is used only as a frame-level reconstruction operation to suppress local temporal fluctuations and does not alter the underlying ordinal likelihood scores. The degradation observed under excessive smoothing is consistent with temporal over-smoothing, which may blur short anomaly responses and their boundaries.

\begin{table}[H]
    \centering
    \caption{\textbf{Effect of Gaussian smoothing on MSAD.}
    Results are frame-level AUC (\%).}
    \label{tab:smoothing_sensitivity}
    \footnotesize
    \setlength{\tabcolsep}{12pt}
    \begin{tabular}{@{}cc@{}}
        \toprule
        \(\sigma\) & AUC (\%) \\
        \midrule
        0  & 88.13 \\
        5  & \textbf{88.18} \\
        10 (default) & 87.55 \\
        20 & 86.65 \\
        \bottomrule
    \end{tabular}
\end{table}

\subsection{Efficiency Analysis}
\label{subsec:efficiency}

We further compare the end-to-end efficiency of Probe-VAD with the complete
caption-mediated pipeline on the same fixed subset of 50 MSAD test videos.
The subset contains 2,235 clips, corresponding to 22,350 sampled input
frames under the ten-frame-per-clip setting. Both pipelines operate on the
same videos and sampled-frame budget.

As shown in Table~\ref{tab:efficiency}, Probe-VAD achieves an effective
throughput of 3.14 FPS, compared with 1.29 FPS for the complete
caption-mediated pipeline, corresponding to a 2.44$\times$ increase in
end-to-end throughput. Probe-VAD additionally reduces persistent
intermediate artifact storage from 6.06 MB to 0.77 MB, an 87.3\%
reduction.

Here, effective FPS is computed from the sampled RGB frames actually
processed by the model rather than all original video frames. The 50-video
subset contains 35,524 original annotated frames, while 22,350 frames are
sampled and provided to the VLM. Transient visual representations are not
counted as persistent storage unless explicitly materialized to disk.

\begin{table}[!t]
    \centering
    \caption{\textbf{End-to-end efficiency on a fixed 50-video MSAD subset.}
    The subset contains 2,235 clips and 22,350 sampled frames. Effective FPS
    is computed from processed frames, while raw videos are excluded from the
    storage comparison.}
    \label{tab:efficiency}
    \footnotesize
    \setlength{\tabcolsep}{4.5pt}
    \begin{tabular}{@{}lccc@{}}
        \toprule
        Pipeline & Time & Effective FPS & Storage \\
        \midrule
        Probe-VAD (10 thresholds)
        & \textbf{1:58:37} & \textbf{3.14} & \textbf{0.77 MB} \\
        Complete caption pipeline
        & 4:49:19 & 1.29 & 6.06 MB \\
        \midrule
        Relative improvement
        & \textbf{59.0\% lower} & \textbf{2.44$\times$} & \textbf{87.3\% lower} \\
        \bottomrule
    \end{tabular}
\end{table}

\begin{figure*}[!t]
    \centering
    \includegraphics[width=0.98\textwidth]{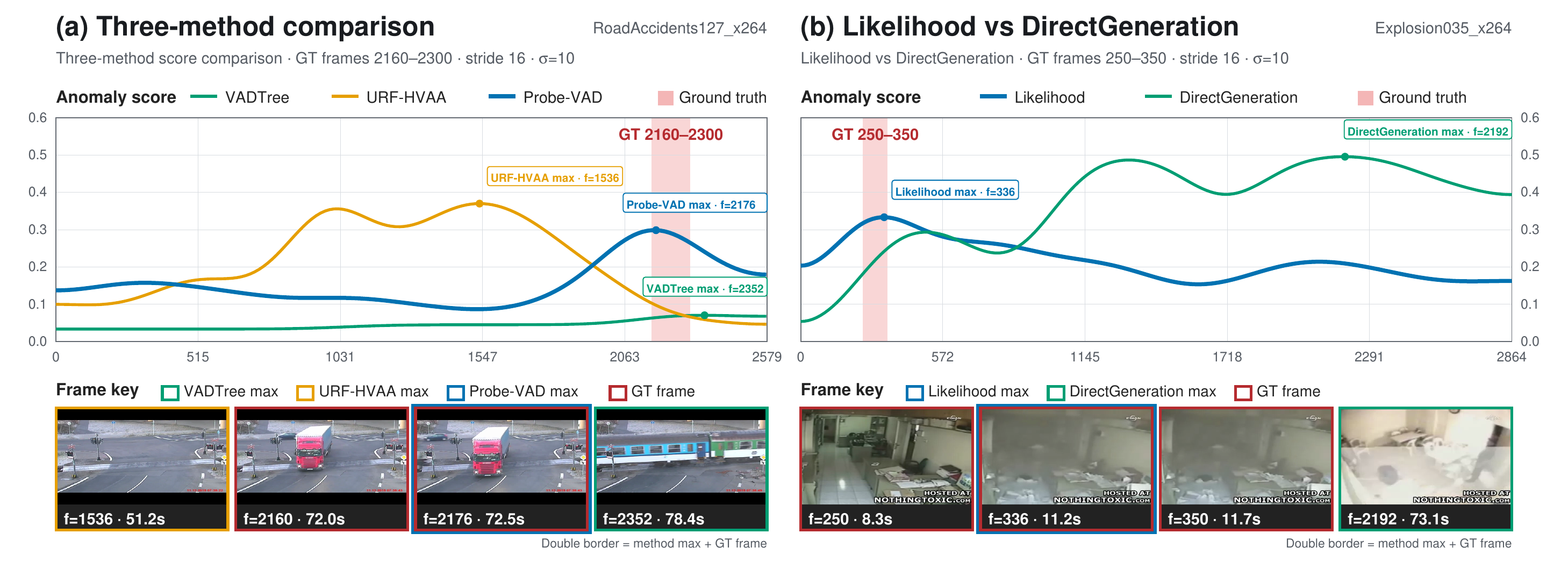}
    \caption{Representative temporal score profiles. (a) Probe-VAD reaches its maximum within the annotated road-accident interval, whereas the compared training-free methods exhibit less aligned temporal responses. (b) Likelihood-based scoring reaches its maximum within the annotated explosion interval, while Direct Generation peaks substantially later.}
    \label{fig:method_compare}
\end{figure*}

\begin{figure*}[!t]
    \centering
    \includegraphics[width=\textwidth,keepaspectratio]{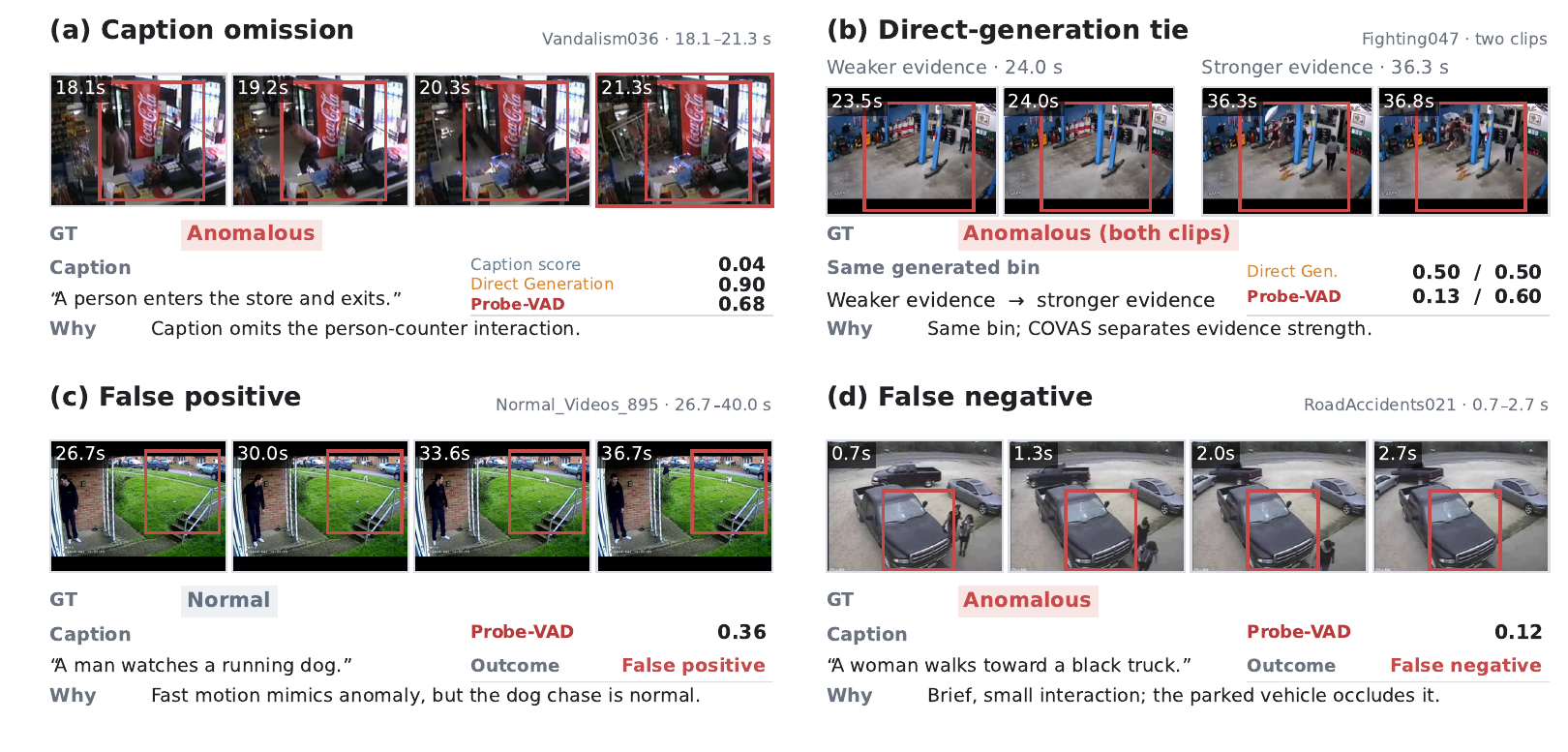}
    \caption{\textbf{Qualitative comparison and representative failure cases}. (a) Caption mediation omits anomaly-relevant interaction cues. (b) Direct numerical generation assigns the same score to visually distinct clips, whereas Probe-VAD separates them through likelihood-level preferences. (c) A false positive associated with salient but benign motion. (d) A false negative involving brief, small, and partially occluded visual evidence.}
    \label{fig:qualitative_failure}
\end{figure*}

Importantly, the storage comparison concerns persistent intermediate artifacts rather than the original visual input or transient runtime representations. Both pipelines use the same source videos and temporal sampling protocol, so raw videos are excluded. Transient visual features are counted only if explicitly materialized to disk.

\section{Discussion and Limitations}
\label{sec:discussion}

\noindent\textbf{Ordinal Consistency}.
The cumulative formulation assumes that support should not increase as the severity threshold becomes more demanding. Although the threshold propositions share the same formulation and differ only in their severity thresholds, their likelihood preferences are not explicitly constrained to satisfy this monotonic relation. We therefore examine how frequently the raw severity profiles violate the expected ordering and how strongly PAVA modifies them.

The maximum observed difference between scalar scores before and after PAVA is only \(4.44\times10^{-16}\), which is consistent with floating-point precision and the sum-preservation property in Eq.~\eqref{eq:pava_sum_invariance}. PAVA therefore plays a structural rather than performance-enhancing role in the current formulation: it converts the raw threshold-wise likelihood preferences into an ordinally consistent profile without altering the scalar ranking statistic used for evaluation.

As shown in Table~\ref{tab:pava_analysis}, 75.33--87.99\% of clips contain at least one adjacent ordering violation before projection, indicating that semantic nesting alone does not guarantee exact monotonicity. Nevertheless, the mean element-wise correction remains small, ranging from 0.002031 to
0.003827, suggesting that most violations are local perturbations rather than large departures from the expected severity order.
\begin{table}[!t]
    \centering
    \caption{Ordinal-consistency diagnostics before PAVA. ``Viol.'' denotes the percentage of clips with any adjacent ordering violation, and \(\delta_{\mathrm{PAVA}}\) denotes the absolute element-wise correction.}
    \label{tab:pava_analysis}
    \scriptsize
    \setlength{\tabcolsep}{3.3pt}
    \begin{tabular}{@{}cccc@{}}
        \toprule
        Dataset & Viol. (\%) & Mean \(\Delta\) & Max \(\Delta\) \\
        \midrule
        UCF & 87.99 & 0.002031 & 0.114160 \\
        MSAD & 75.33 & 0.002525 & 0.123895 \\
        XD & 80.00 & 0.003827 & 0.235396 \\
        \bottomrule
    \end{tabular}
\end{table}

\noindent\textbf{Qualitative Analysis}. Fig.~\ref{fig:method_compare} presents representative temporal score profiles complementing the controlled comparisons. In the road-accident example, Probe-VAD reaches its strongest response within the annotated anomaly interval, while compared training-free methods exhibit less aligned temporal responses. In the explosion example, likelihood-based scoring peaks within the annotated interval, whereas Direct Generation peaks substantially later. Fig.~\ref{fig:qualitative_failure} further illustrates the two score-extraction bottlenecks and representative failure cases. Caption mediation may omit anomaly-relevant interaction cues, while visually distinct clips can receive the same decoded severity despite remaining distinguishable at the likelihood level. The false-positive and false-negative cases instead expose limitations outside the score-extraction interface, including salient but benign motion and brief, small, or partially occluded anomaly evidence. These examples are intended as qualitative illustrations rather than aggregate evidence.

\noindent\textbf{Limitation}.One limitation of our method is that it inherits perception errors from the frozen VLM, which cannot be resolved through score extraction alone. This is a common limitation for training-free VLM-based approaches. Such perception errors may include mistaking salient but benign motion for anomalous activity, or failing to recognize brief abnormal interactions, small anomaly-related objects, and partially occluded events. These errors may be more pronounced in crowded scenes, low-resolution footage, or ambiguous events. In these cases, Probe-VAD can better preserve and organize the evidence perceived by the VLM, but cannot recover visual evidence that the underlying model itself fails to capture.

\balance
\section{Conclusion}
\label{sec:conclusion}

We introduced Probe-VAD, a training-free framework that translates the visual evidence of a frozen VLM into continuous anomaly-ranking scores through direct visual conditioning and ordinal likelihood probing. Rather than compressing visual observations into captions or relying on a single decoded numerical response, Probe-VAD retains likelihood-level preferences over ordered severity propositions and integrates the resulting cumulative evidence into a continuous ranking statistic. Experiments across UCF-Crime, MSAD, and XD-Violence show that direct visual conditioning consistently outperforms matched caption input, likelihood-based scoring substantially improves over direct numerical generation, and the cumulative binary-threshold interface further improves over flat 10-class likelihood scoring.
Probe-VAD achieves frame-level AUC values of 86.27\%, 87.55\%, and 92.11\% on UCF-Crime, MSAD, and XD-Violence, respectively. Experiments with Qwen3-VL-8B-Instruct further show that the proposed score-extraction interface is not tied to a single frozen VLM backbone. Together, these results demonstrate that retaining direct visual evidence,
preserving likelihood-level preferences, and organizing them according to ordinal severity structure provide an effective interface for converting pretrained VLM understanding into continuous temporal anomaly rankings.
\FloatBarrier
\balance

\bibliography{citation}
\bibliographystyle{IEEEtran}

\end{document}